\documentclass[letterpaper, 10 pt, conference]{ieeeconf}  

\IEEEoverridecommandlockouts                              

\usepackage{graphicx} 
\usepackage{amsmath} 
\usepackage{amssymb}  
\usepackage[ruled,linesnumbered]{algorithm2e}
\usepackage{bm}
\usepackage{booktabs}
\usepackage{threeparttable}
\usepackage{cite}
\usepackage{multirow}
\usepackage{multicol}
\usepackage{backref}
\title{\LARGE \bf
Optimizing Facial Expressions of an Android Robot Effectively: a Bayesian Optimization Approach
}

\author{Dongsheng Yang$^{1,2}$, Wataru Sato$^{2,3}$, Qianying Liu$^{1}$, Takashi Minato$^{5}$, Shushi Namba$^2$,  Shin'ya Nishida$^{1,4}$
\thanks{$^{1}$Graduate School of Informatics, Kyoto University, Kyoto, Japan
    {\tt\small yang.dongsheng.46w@st.kyoto-u.ac.jp}, $^{2}$Psychological Process Research Team, Guardian Robot Project, RIKEN, Kyoto, Japan, $^{3}$Field Science Education and Research Center, Kyoto University, Kyoto, Japan, $^{4}$NTT Communication Science Laboratories, Nippon Telegraph and Telephone Corporation, Atsugi, Japan and $^{5}$Interactive Robot Research Team, Guardian Robot Project, RIKEN, Kyoto, Japan}%
    \thanks{This work is supported by JSP.}
}

\begin{document}

\maketitle
\thispagestyle{empty}
\pagestyle{empty}

\begin{abstract}

Expressing various facial emotions is an important social ability for efficient communication between humans. A key challenge in human-robot interaction research is providing androids with the ability to make various human-like facial expressions for efficient communication with humans. The android Nikola, we have developed, is equipped with many actuators for facial muscle control. While this enables Nikola to simulate various human expressions, it also complicates identification of the optimal parameters for producing desired expressions. Here, we propose a novel method that automatically optimizes the facial expressions of our android. We use a machine vision algorithm to evaluate the magnitudes of seven basic emotions, and employ the Bayesian Optimization algorithm to identify the parameters that produce the most convincing facial expressions. Evaluations by naïve human participants demonstrate that our method improves the rated strength of the android’s facial expressions of anger, disgust, sadness, and surprise compared with the previous method that relied on Ekman’s theory and parameter adjustments by a human expert.

\end{abstract}

\section{Introduction}\label{sec:intro}


Robot facial expressions (RFEs) have attracted attention in the field of human-robot interactions because facial expressions are essential for human emotional communication in everyday life. Leite et al. \cite{leite2013social} suggested that robot expressions can substantially influence human-robot interactions because they attract attention from humans and generate long-term memories.



\begin{figure}[t]
  \centering
  \includegraphics[width=0.45\textwidth]{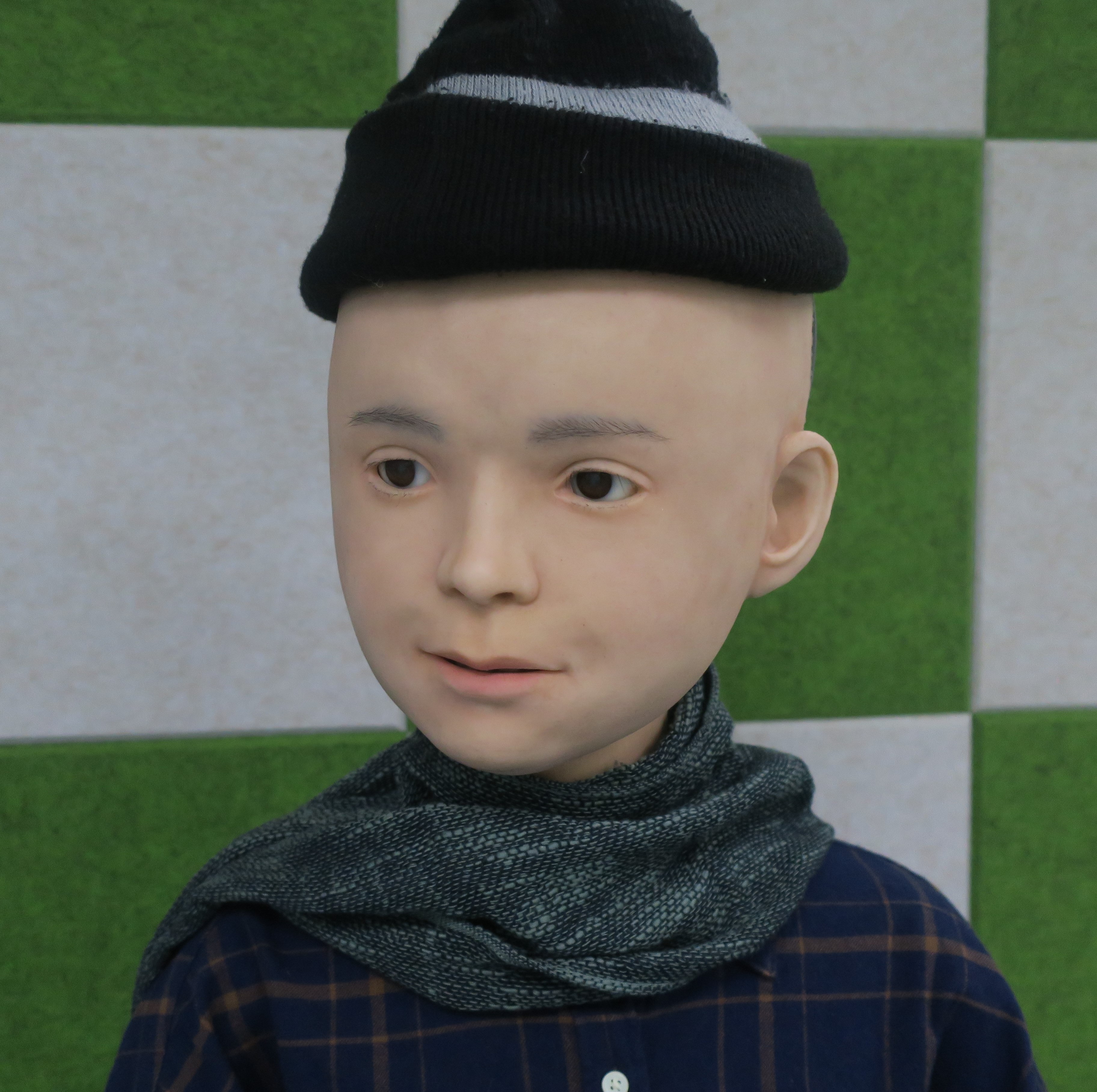}
  \caption{Nikola, an android developed by RIKEN.}
  \label{fig:intron}
\end{figure}

Giving many action parameters comparable to human facial movements is a reasonable hardware strategy for enabling androids to express rich human-like emotions. A significant number of androids have been developed to reproduce human-like facial expressions (see \cite{kobayashi1994study,berns2006control,hashimoto2006development,oh2006design,hashimoto2008dynamic} and Table \ref{tab:robotcomparison}). The android Nikola (Fig. \ref{fig:intron}), we have developed, is one of the world’s leading androids in this respect. Through a series of human experiments, Sato et al. \cite{sato_android_2022} verified that Nikola can produce several basic human-like RFEs. 


Nikola is an android with the face of a human child; it was designed for use in studies of interactions between humans and robots, with a focus on emotional communications. Soft and deformable silicone skin is attached to the front of Nikola’s skull. Beneath the skin, there are 35 pneumatic actuators with pressure control valves: 29 for facial muscle movement control, 3 for head movement control (roll, pitch, and yaw rotation), and 3 for eyeball control (i.e., panning of the individual eyeballs and tilting of both eyeballs)  \cite{sato_android_2022}.


However, when an android has many controllable parameters, it becomes difficult to identify the optimal parameter set for a particular goal. Recent studies have enabled robots to generate facial expressions, either through knowledge-based human coding or in an automated manner\cite{rawal2022facial}. Sato et al. \cite{sato_android_2022} used the knowledge-based human coding approach. They identified basic axes for each emotion based on Ekman’s theory \cite{facs2002}, and then asked a human expert to adjust the parameters of each axis. This approach generated convincing positive expressions but did not yield robust negative expressions.

Here, we attempted to improve expressions (i.e., generate emotional expressions that are more convincing to human observers) by controlling more parameters; however, this approach is both challenging and time-consuming for human operators. Therefore, instead of using a human coding approach, we used an automated approach.

\begin{table*}[h]
    \centering
    \caption{Comparison of Nikola with other facially expressive humanoid robots.}
     \begin{threeparttable}
    \begin{tabular}{ccccc}
        \toprule
        Android* & Head DOF** & Head Size & Emotional expression capability & Validation \\
        \midrule
        Nikola (this study) & 35 & Child & 7 emotions & Emotion recognition  \\
        Face robot\cite{kobayashi1994study} & 24 & Adult  & 6 basic emotions & Emotion recognition (no statistical test)\cr
        ROMAN\cite{berns2006control} & 21 & Adult& 6 basic emotions & Emotion recognition\cr
        Saya\cite{hashimoto2006development} & 24 & Adult & 6 basic emotions & Emotion recognition (no statistical test)\cr
        Albert HUBO\cite{oh2006design} & 31 & Adult & Full range (not specified) & - \cr
        Face Robot\cite{hashimoto2008dynamic} & 39 & Adult & 6 basic emotions & - \cr
        Geminoid F\cite{becker2011intercultural} & 12 & Adult & 5 basic emotions & Emotion recognition\cr
        FACE\cite{mazzei2012hefes} & 32 & Adult & 6 basic emotions & Emotion recognition (no statistical test) \cr
        Face robot \cite{lin2016expressional} & 7 & Child & 6 basic emotions & Emotion recognition (no statistical test) \cr
        EveR \cite{hyung2019optimizing} & 16 & Adult & 6 basic emotions & - \\
        XIN-REN \cite{ren2016automatic} & 7 & Adult & - & - \\
        EVA \cite{faraj2021facially} & 25 & Adult & 6 basic emotions & - \\
        EVA 2.0 \cite{chen2021smile} & 25 & Adult & - & - \\
        \bottomrule
    \end{tabular}
    \begin{tablenotes}
        \item[*]  List only includes androids with human-like appearance
        \item[**] DOF = degrees of freedom
    \end{tablenotes}
\end{threeparttable}
    \label{tab:robotcomparison}
\end{table*}

Automatic RFE optimization is challenging in two respects. First, RFE performance must be evaluated both reliably and automatically. Because Nikola’s face is reasonably human-like, we used Py-Feat \cite{cheong2021py}, an open-source human emotion recognition model, for automatic evaluation of RFEs. Second, the complexity of parameter optimization increases exponentially with increasing degrees of freedom (DOF). Nikola has 35 actuators on its face. Thus, Nikola has a greater number of DOF that require optimization than most other androids (Table \ref{tab:robotcomparison}). 

In a related study concerning the automatic generation of RFEs, Park \cite{park2015generation} proposed using a dynamic emotion model (based on the linear affect-expression space model) to generate RFEs. However, because this method does not consider the nonlinear feature of most physical robots, it is difficult to use them. 
WE-4R \cite{itoh2006mechanical} used emotion vectors for facial expression generation. Ren et al.\cite{ren2016automatic} proposed a kinematics-based learning method for robot XIN-REN to imitate expression of a human subject. 
Hyung et al.\cite{hyung2019optimizing} used a genetic algorithm to identify optimal facial expressions in the search space. Similarly, Habib \cite{habib2014learning} proposed a learning-based method with a genetic algorithm capable of inverse nonlinear mapping from human faces to robot actuators. However, a recent study suggested that the increasing complexity of robots (i.e., increased number of DOF) is leading to difficulty in the application of genetic algorithm methods \cite{chen2021smile}. Rawal \cite{rawal2022exgennet} proposed a deep generative method for generating emotions, in which grid search and gradient descent are combined to automatically identify the parameters of desired RFEs. Importantly, this learning-based method is potentially data-intensive.

We consider this problem to be equivalent to an expensive black-box optimization issue \cite{jones1998efficient} for an entire system. Therefore, we propose the use of Bayesian Optimization \cite{pelikan1999boa} to identify a set of optimal parameters that enable Nikola to generate higher-quality facial expressions. We call this method Bayesian optimization-based robot facial expression optimization (\textbf{BORFEO}). In summary, our contributions are threefold:

\begin{enumerate}
    \item We develop a method, BORFEO, which automatically identifies a set of optimal facial parameters that allow Nikola to generate prototypical facial expressions without initialization.
    \item BORFEO converges within a relatively small number of trials, which suggests that BORFEO is an efficient RFE optimization procedure.
    \item BORFEO can generate RFEs that receive higher human ratings for most basic emotions compared with previous work involving the same robot \cite{sato_android_2022}; this finding indicates that BORFEO is a high-performance method.
\end{enumerate}


The remainder of this paper is structured as follows. Section \ref{sec:preli} describes the preliminary setup. Section \ref{sec:borfeo} describes the BORFEO methodology. Section \ref{sec:exp} describes the experiments conducted to generate Nikola’s RFEs and evaluate their performance. Section \ref{sec:resultanddiscussion} presents the experimental results and discussion. Section \ref{sec:conclude} presents the conclusions and offers suggestions for future work.




\section{Preliminary setup}\label{sec:preli}

\subsection{Prototypical RFEs in Nikola}

Nikola’s actuator arrangement design is based on the facial action coding system (FACS) proposed by Ekman \cite{facs2002}. A certified FACS coder conducted the action unit mapping from actuators to action units, as defined by Ekman \cite{facs2002}; this mapping is also the basis of human-coded prototypical RFEs. The detailed correspondence of FACS coding is shown in Table \ref{tab:robotactuators}. Further details regarding the differences between Nikola and other androids are described in our previous work\cite{sato_android_2022}.




\subsection{Emotion recognition}

Py-Feat \cite{cheong2021py} is a Python-based open-source facial expression analysis toolbox developed by Cheong and colleagues at Dartmouth College; it includes state-of-the-art facial expression models. Py-Feat can be used for tasks such as face detection, facial landmark detection and alignment, face or head pose estimation, action unit detection, and emotion detection.

For emotion detection, Py-Feat provides four models: a random forest model trained on the histogram of oriented gradients extracted from ExpW, CK+, and JAFFE datasets; a support vector machine model trained on the histogram of oriented gradients extracted from the same dataset; a deep convolution network; and ResMaskNet (a residual masking network for recognition of facial expressions) \cite{pham2021facial}. We chose the ResMaskNet model because its performance was superior to the other three models in the emotion recognition task using the FER2013 dataset.




\subsection{Experimental setup}
\begin{figure}[htpb]
  \centering
  \includegraphics[width=0.45\textwidth]{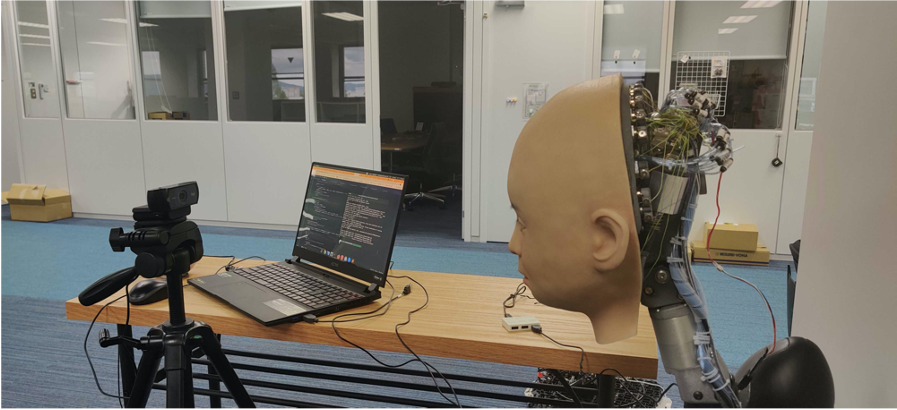}
  \caption{Video recording setup.}
  \label{fig:rc_system}
\end{figure}

Figure \ref{fig:rc_system} shows the experimental setup used for data collection and assessment of our proposed algorithms. We used a Logicool C922 Pro stream web camera (HD1080P) to capture RGB images. For robot control and camera manipulation, we employed a notebook computer (Intel Core i7-10870H CPU, 2.20 GHz, NVIDIA GeForce RTX 3060 laptop GPU, GDDR6 6GB) running the Ubuntu 20.04 system. Nikola uses ROS (Robot Operating System) Noetic Ninjemys as its base system; this software was released on May 23, 2020 and is compatible with Ubuntu 20.04. We connected a web camera to the laptop through a USB cable and connected Nikola to the laptop via Wi-Fi. The position of the web camera was fixed throughout the experiment.


\begin{figure}[htpb]
  \centering
  \includegraphics[width=0.45\textwidth]{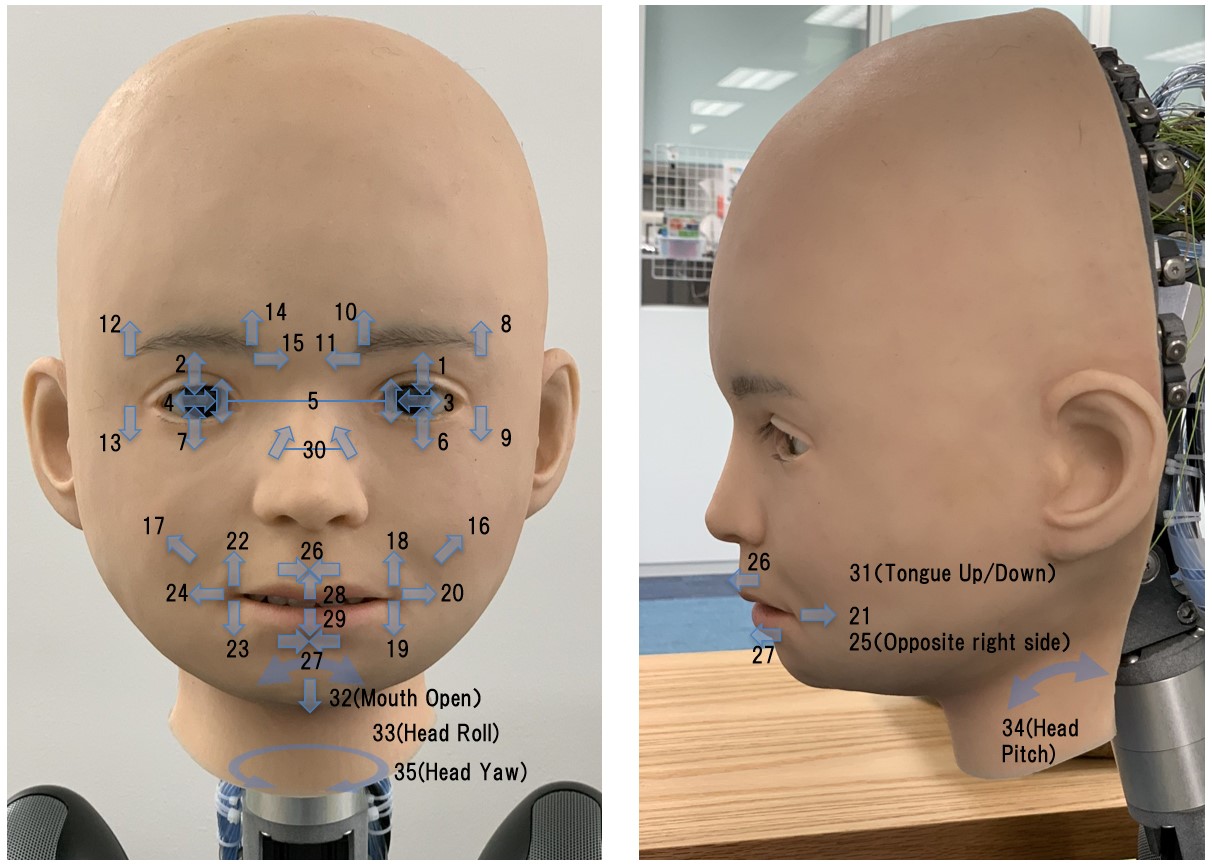}
  \caption{Actuator arrangement in Nikola.}
  \label{fig:nikola_actuators}
\end{figure}

\begin{table}[h]
    \centering
    \caption{Actuator details.}
     \begin{threeparttable}
    \begin{tabular}{lcc}
        \toprule
        \textbf{Actuator} & Description & Side\\
        \midrule
        1&Upper lid raiser&L\\
        2&Upper lid raiser&R\\
        6&Cheek raiser/Lid tightener&L\\
        7&Cheek raiser/Lid tightener&R\\
        8&Outer brow raiser&L\\
        9&Cheek raiser&L\\
        10&Inner brow raiser&L\\
        11&Brow lowerer&L\\
        12&Outer brow raiser&R\\
        13&Cheek raiser&R\\
        14&Inner brow raiser&R\\
        15&Brow lowerer&R\\
        16&Cheek puller&L\\
        17&Cheek puller&R\\
        18&Lip corner puller&L\\
        19&Lip corner depressor&L\\
        20&Lip stretcher&L\\
        22&Lip corner puller&R\\
        23&Lip corner depressor&R\\
        24&Lip stretcher&R\\
        28&Lip funneler&T\\
        29&Lip funneler&B\\
        30&Nose wrinkler&-\\
        32&Jaw dropper&-\\
\bottomrule
    \end{tabular}
    \begin{tablenotes}
        \item[*] L denotes left side, R denotes right side, T denotes top, B denotes bottom, and - denotes not applicable.
    \end{tablenotes}
\end{threeparttable}
    \label{tab:robotactuators}
\end{table}

\section{Methodology}\label{sec:borfeo}
Nikola’s action unit is controlled by a pneumatic actuator with control inputs that range from 0 to 255 (Fig. \ref{fig:nikola_actuators}). To optimize Nikola’s facial expressions, we transformed the task into a black-box function optimization problem and proposed the BORFEO method. Because Py-Feat can evaluate a robot’s facial expression and generate probability scores for seven basic emotions (anger, disgust, fear, happiness, sadness, surprise, and neutral), we regarded the parameters of the control axes as the input of the black-box function; the Py-Feat probability scores were the black-box output. We sought to identify a set of values for the control axes parameters allowing Nikola’s facial expressions to achieve high confidence scores on Py-Feat.


There are various methods to address the black-box optimization problem. Grid search and random sampling are used for small-scale systems but are not appropriate for this task because of the very large parameter space. Compared with other algorithmic-based methods such as the genetic algorithm, Bayesian Optimization achieves rapid convergence, and its performance has proven superior to other methods for tasks such as the 2020 Black-Box Optimization Challenge \cite{turner2021bayesian}, hyperparameter optimization \cite{alibrahim2021hyperparameter}, and nutrition problems \cite{gumustekin2014comparative}. Thus, we use Bayesian Optimization to solve the black-box function optimization problem.

In this section, we introduce the task formulation and then describe the use of Bayesian optimization to identify facial expressions.

\subsection{Task formulation}

Formally, given $n=35$ parameters of control axes of the robot $x=(a_0, ..., a_n)$, which satisfies $0 \leq a_i \leq 255, a_i \in \mathbb{Z}$, we define the parameter space as $\mathcal{X}$. Each set of parameters $x$ can generate a corresponding facial expression, which is assigned a score $y$ by Py-Feat. We define the process of obtaining the Py-Feat score $y$ from a set of control axes parameters $x$ as $y=f(x)$. The goal is to identify an optimal set of x that satisfies the following condition:

\begin{equation}
x^* = {argmax}_{x \in \mathbb{X}} f(x) 
\end{equation}

$f(x)$ is an expensive black-box function, for which no analytical description or gradient information exists concerning the objective function.

\subsection{Bayesian Optimization}

The main objective of the algorithm is to update a Bayesian statistical model of the black-box function and use an acquisition function to determine the next search point. We first generate an initial set of candidate solutions and then find the next possible optimal data point based on these solutions points, add that point to the candidate solutions set, and repeat this step until the end of the iteration. 
Finally, the optimal data point in the set of candidate solutions is taken as the solution to the problem.

However, $f(x)$ is a black-box function, thus the key issue here is how to determine the next search point based on the already searched points. 
As we show in Algorithm \ref{alg:boa}, the Bayesian optimization algorithm uses Gaussian Process (GP) $\mathcal{M}$ to model the black-box function. For each $x \in \mathbb{X}$ we can obtain the mean $\mu_x$ and variance $\delta_x$. Given {$(\mu, \delta)$} pairs, we can evaluate each point with the \textit{Acquisition Function} $\mathcal{S}$ to obtain the next search point. 

\begin{algorithm}[!h]
\SetAlgoLined
\KwIn{$f, \mathbb{X}, \mathcal{M}, \mathcal{S}$}
\KwOut{$x^*$}
 initialize $(f, \mathbb{X}) \mapsto \mathcal{D}$\;
\For{\texttt{$\mathcal{D} \mapsto i$} to T}{
\texttt{FITMODEL($\mathcal{M},\mathcal{D}$)} $\mapsto$ $p(y|x,\mathcal{D})$\; 
$argmax_{x \in \mathbb{X}S(x,p(y|x,\mathcal{D}))}$\;
$f(x_i) \mapsto y_i$\;
$\mathcal{D} \cup (x_i,y_i) \mapsto \mathcal{D}$\;
}
\caption{Bayesian Optimization.}
\label{alg:boa}
\end{algorithm}

Where $\mathcal{D}$ is defined as a dataset that consists of several pairs of data, each represented as $(x_i,y_i)$; $x_i$ is a set of hyperparameters, and $y_i$ represents the result corresponding to the set of hyperparameters.

\subsubsection{Gaussian Process}\label{sec:gp}

Consider a sequence of continuous random variables $\{x_i\}$, if the vector formed by any subsequence $[x_{t_0},...,x_{t_k}]^\intercal$ obeys a multidimensional normal distribution, this sequence of random variables is defined as a GP. 

Notably, our variables $x$ are discrete, while GP requires continuous random variables. We ignore this discontinuity here because the parameters $a_i$ of $x$ are continuous integers that can be regarded as approximately continuous. 


Specifically, consider $k$ random variables $x_1,...,x_k$ that follow a k-dimensional Gaussian distribution:

\[
     N(\bm{\mu}_k,\bm{\Sigma}_k)
\]

where the mean vector satisfies $\bm{\mu}_k\in\mathbb{R}^k$ and the covariance matrix satisfies $\bm{\Sigma}_k\in\mathbb{R}^{k \times k}$. 

Because the integral of the normal distribution yields an analytic solution, the marginal and conditional probabilities can easily be obtained. Thus, given a set of control axis parameter values $\bm{x}_i,i=1,...,l$ and the corresponding set of scores $y$, we can model the black-box function $y=f(x)$ from these samples. Given an input $x$, we can use the function to predict the label value $y$ and posterior probability $p(y|\bm{x})$. This process is referred to as \textit{Gaussian Process Regression} (GPR).

GPR can use a set of sampled solution values to fit a Bayesian model and produce the probability distribution of the solution values. Given the black-box function $f(x)$ that satisfies the following mapping:
\[
    \mathbb{R}^n \rightarrow \mathbb{R}
\]
GPR will update the model based on $x_i, i=1,...,t$ and the corresponding  solution $f(x_i)$ to fit the black-box function. Indeed, the model describes the probability distribution of $f(x)$.

\subsubsection{Acquisition Function}\label{sec:acquifunc}

The acquisition function is an inexpensive function that can be evaluated at a given point that guides how the parameter space should be explored for the optimization problem. The acquisition function can then be optimized to select the data points for the next observation. This approach replaces the original optimization problem with another function $a(x)$, which can be optimized more easily.

In our study, we use the Upper Confidence Bound (UCB) acquisition function, which is defined as:

\begin{equation}
    a_{UCB}(x;\beta) = \mu(x) - \beta \delta(x)
\end{equation}

The terms $\mu(x)$ and $\delta(x)$ can be interpreted as explicitly encoding a trade-off between exploitation (evaluating at points with low mean) and exploration (evaluating at points with high uncertainty). $\beta > 0$ is a trade-off hyperparameter.



\section{Experiment}\label{sec:exp}

\begin{figure*}[t]
  \centering
  \includegraphics[width=0.95\textwidth]{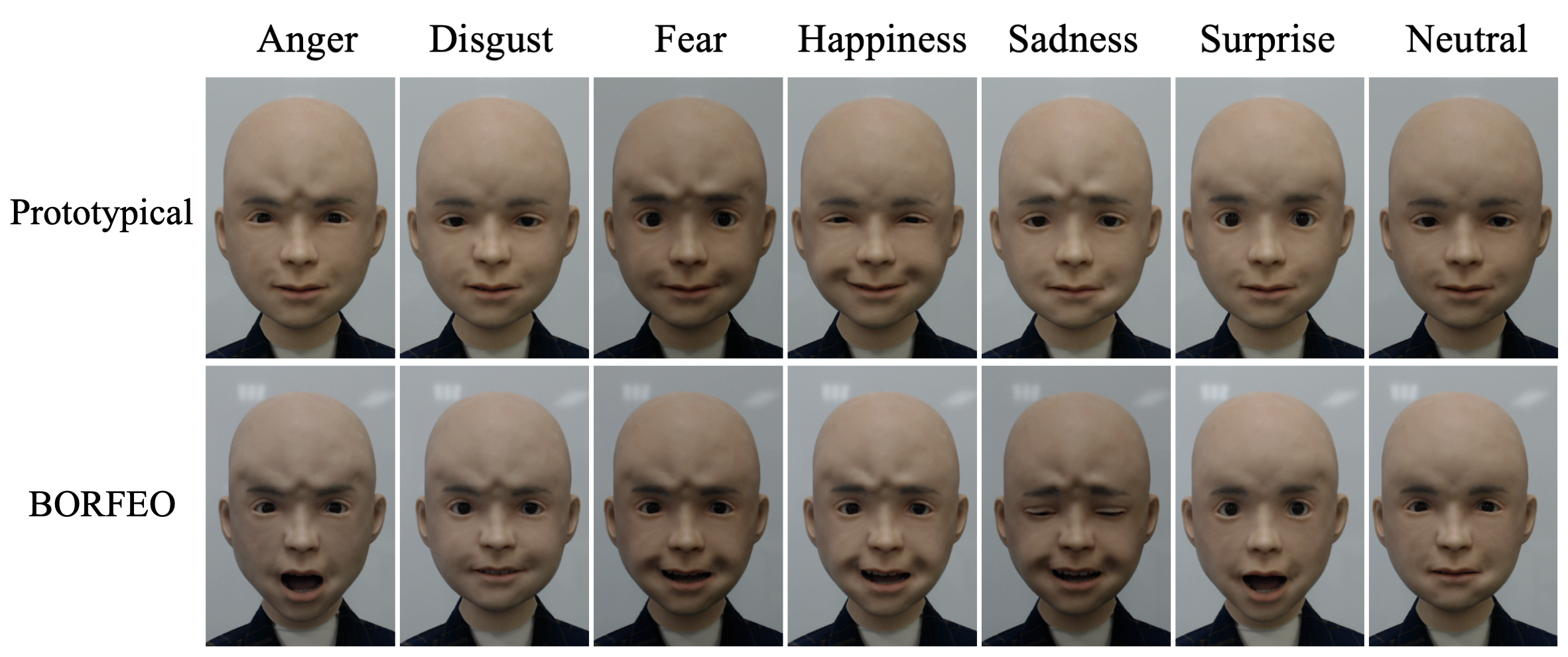}
  \caption{Illustrations of prototypical and BORFEO-generated RFEs produced by the android Nikola.}
  \label{fig:exampleforprovsborfeo}
\end{figure*}

\subsection{Actuator Selection}
With some prior knowledge, we restricted the movements of actuators in optimization.


\begin{enumerate}
    \item \textbf{Safety}. Because of Nikola’s mechanical limitations, several pairs of fragile points required careful adjustment to random values. For example, actuators ${s_{18}}$ and $s_{19}$ constitute a pair of fragile points because they will pull the same part in opposite directions.
    \item \textbf{Psychological knowledge}. Although some psychological evidence suggests that gaze/head direction is involved in emotional expression \cite{semyonov2021beyond}, the gaze/head variant is not implemented in most prototypical facial expression databases or Py-Feat. Therefore, we discarded this factor. 
    \item \textbf{Symmetric}. Nikola’s actuator arrangement is symmetrical on the left and right. Considering that the seven basic facial expressions are symmetrical \cite{ekman1978facial}, we simplify the problem and assign the same value to all symmetrical pairs of actuators. $a_i = a_j$, which satisfies $0 \leq i, j \leq 35$, where $i$ and $j$ are center-symmetrical actuators.
\end{enumerate}

Considering these points, we selected 24 actuators with a total dimension equal to 14, as shown in Table \ref{tab:robotactuators}.

\subsection{RFE generation}
The BORFEO system was written in Python 3.7.8. We used FFmpeg to control the web camera and Bayesian Optimization library \cite{bigye2014bo} for implementing BORFEO. ROS serves as the bridge system to send instructions to Nikola\cite{ros}. To evaluate our method, we conducted a 100-round BORFEO and generated 100 RFEs for each of the seven basic emotions. The number of rounds was determined by the preliminary experiment. Since we didn't optimize the processing time, each round of BORFEO took about 20 seconds including 10 seconds for image capturing by a webcam and 5 seconds for Py-Feat processing. Since we carried out experiments online, each session took about 35 minutes. \ref{fig:exampleforprovsborfeo} shows the top-scoring BORFEO-generated expressions, together with the human-coded (mouth closed) prototypical expressions \cite{sato_android_2022}. 

\begin{figure}[htpb]
  \centering
  \includegraphics[width=0.45\textwidth]{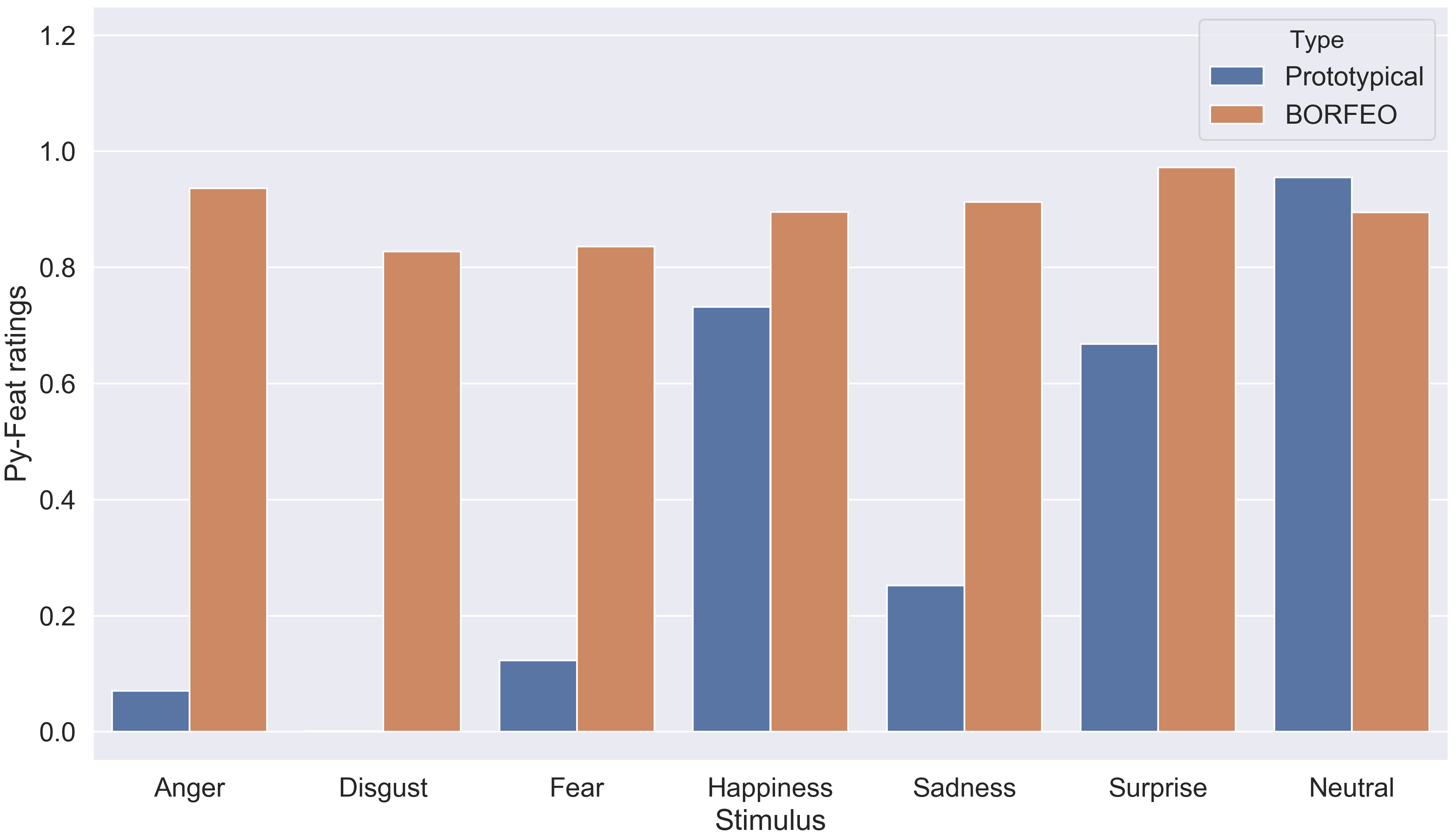}
  \caption{Py-Feat rating results.}
  \label{fig:pyfeatres}
\end{figure}

Figure \ref{fig:pyfeatres} shows the machine rating results. Clear improvement in the Py-Feat rating was observed for all basic emotions (except neutral), which implies that BORFEO can identify a set of RFEs that receive higher Py-Feat ratings compared with prototypical RFEs.

\subsection{Human evaluation}

To evaluate differences in human observers’ ratings between human-coded prototype RFEs and optimized RFEs, we recruited 40 naïve Japanese participants (20 men, 20 women; mean ± standard deviation age = 33.7 ± 5.1 years) in an online survey experiment.

To facilitate correspondence with Py-Feat ratings, we designed a more comprehensive survey compared with most previous studies (see \cite{kobayashi1994study,berns2006control,hashimoto2006development,becker2011intercultural,mazzei2012hefes,lin2016expressional}). Participants were given stimulus photos of prototypical and BORFEO-generated expressions for seven basic emotions; they were asked to rate the extent to which each photo depicted one of the seven basic emotions using a 7-point scale ranging from 1 (does not depict that emotion) to 7 (strongly depicts that emotion). The prototypical expressions were generated using the method described in \cite{sato_android_2022}, based on Ekman’s theory and parameter adjustments by a human expert. To increase stimulus variability, we added mouth-open conditions to the original mouth-closed conditions. We selected the five best (but dissimilar) BORFEO-generated expressions for each emotion. The orders of stimulus photos and rated emotions were randomized. The human evaluation experiments were conducted via Qualtrics (an online questionnaire system) using a procedure approved by the ethical committee of RIKEN.






\section{Results and Discussion}\label{sec:resultanddiscussion}
\begin{table*}[htpb] 
    \centering 
    \caption{Result of two-way repeated measures analysis of variance.}
    \begin{threeparttable} 
    \begin{tabular}{cccccccc}  
        \toprule[1.5pt]         
        Source\tnote{a} & SS\tnote{b} & NDF\tnote{c} & DDF\tnote{d} & Mean squares & F-value & p-unc\tnote{e} & $\eta^2_p$ \tnote{f} \cr
        \midrule 
        RFE     & 0.763050  & 1 & 39 & 0.763  & 12.52  & $<$0.0001 & 0.2430 \cr
        Emotion   & 40.695060  & 6 & 234 & 6.78  & 41.35  & $<$0.0001 & 0.5146	 \cr
        RFE * Emotion  & 3.147407 & 6 & 234 & 0.524 & 8.18 & $<$0.0001  & 0.1734 \cr
        \bottomrule[1.25pt] 
    \end{tabular}
    \begin{tablenotes}
        \item[a] ``Source" column shows groups of RFEs (3 RFEs per group). 
        \item[b] SS = sums of squares
        \item[c] NDF = degrees of freedom (numerator)
        \item[d] DDF = degrees of freedom (denominator)
        \item[e] p-unc: uncorrected p-value 
        \item[f] $\eta^2_p$: partial eta-square effect size
    \end{tablenotes}
     
    \label{tab:twowayrepeatedanova}
    \end{threeparttable}
\end{table*}

\subsection{Target emotion recognition}

\begin{figure}[htpb]
  \centering
  \includegraphics[width=0.45\textwidth]{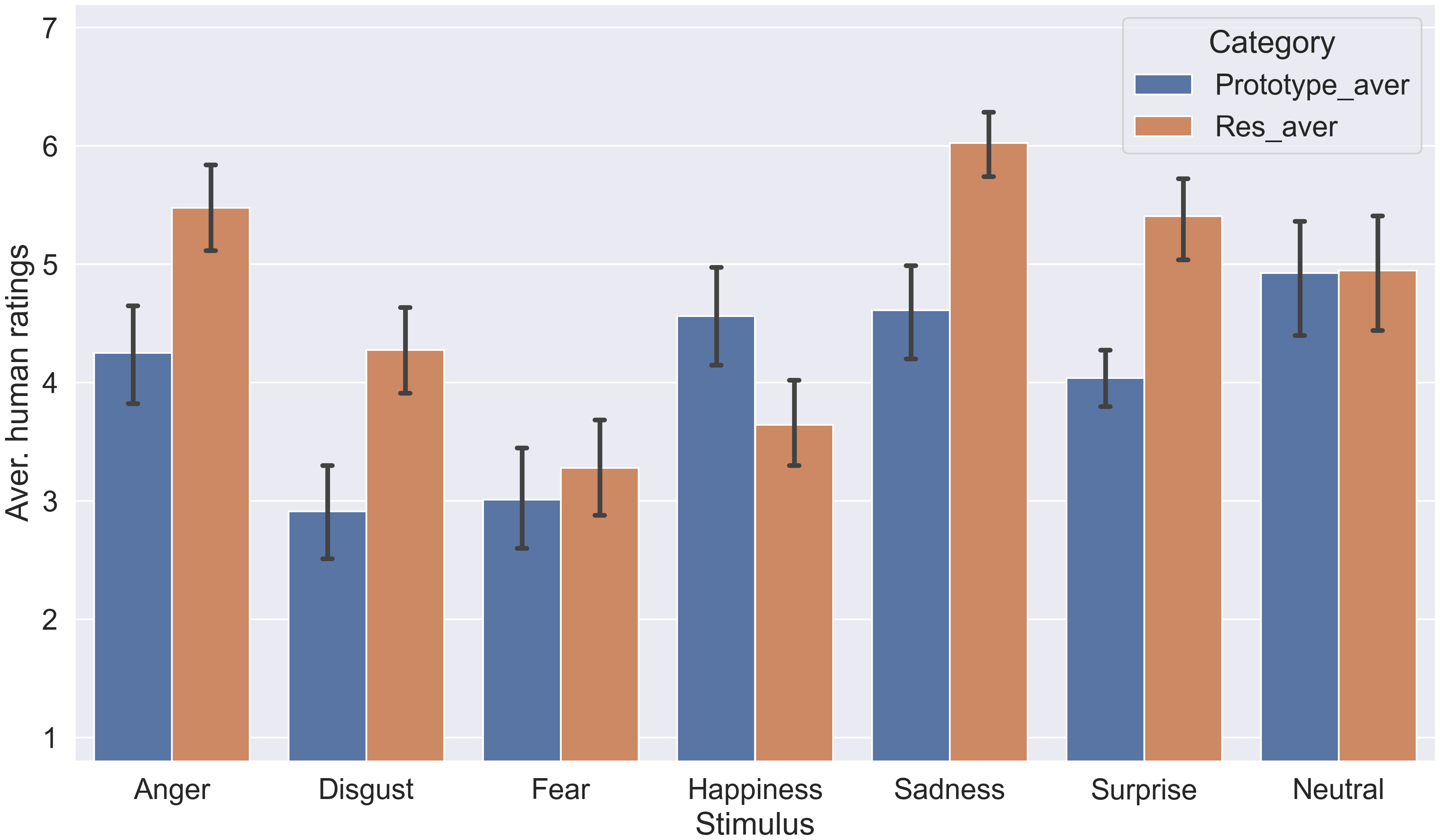}
  \caption{Main results of the human evaluation experiment.}
  \label{fig:mainres}
\end{figure}

Figure \ref{fig:mainres} shows the mean target emotion ratings with 95\% confidence intervals for the seven basic emotions and two types of RFEs. The target emotion ratings were analyzed using two-way repeated measures analysis of variance with RFE (prototypical and BORFEO) and emotion (anger, disgust, fear, happiness, sadness, surprise, and neutral) as independent variables (Table \ref{tab:twowayrepeatedanova}). The results showed significant main effects of RFE and emotion, as well as a significant interaction between these factors.


Follow-up analyses for the simple main effects of RFE revealed that, compared with prototypical expressions, BORFEO-generated expressions had significantly higher target emotion recognition scores for angry, disgusted, sad, and surprised expressions (p $< 0.001$). In contrast, prototypical happy expressions had significantly higher target emotion ratings than BORFEO-generated happy expressions (p $< 0.001$).

\begin{figure}[htpb]
  \centering
  \includegraphics[width=0.45\textwidth]{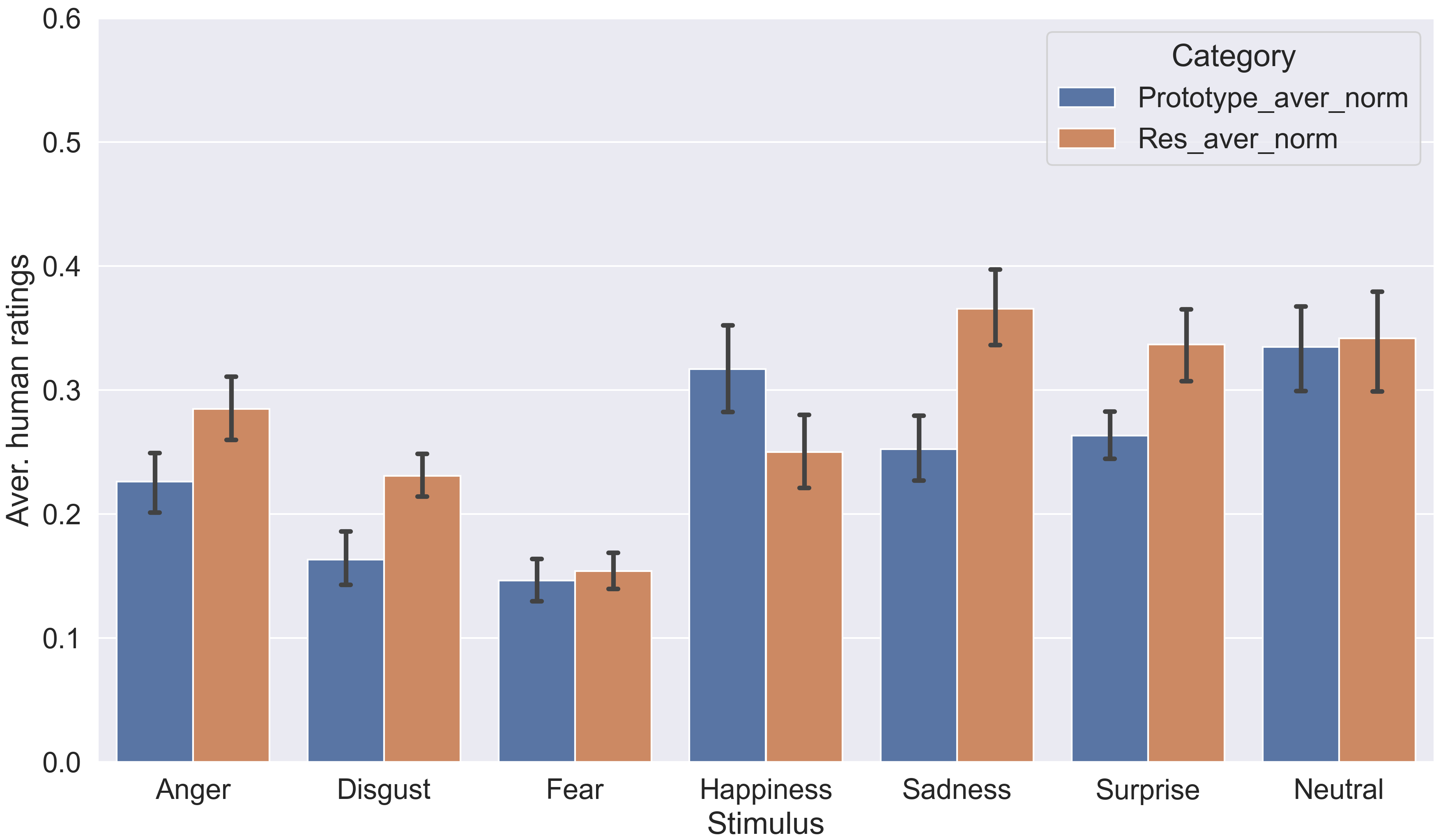}
  \caption{Normalized ratings in the human evaluation experiment.}
  \label{fig:mainres_norm}
\end{figure}

The results in Figure \ref{fig:mainres} do not exclude the possibility that the BORFEO method increases human ratings for both target (e.g., anger) and non-target expressions (e.g., disgust). If such an effect is present, we may be unable to determine whether the BORFEO method produces a more convincing target expression. To address this concern, we also analyzed the normalized target emotion rating, i.e., the target emotion rating divided by the sum of all emotion ratings for the same face image (Fig. \ref{fig:mainres_norm}). We found that the effects of BORFEO were very similar when the analysis focused on raw (Fig. \ref{fig:mainres}) and normalized ratings (Fig. \ref{fig:mainres_norm}), thus eliminating the above concern.

Our results indicate that, compared with prototypical RFEs, the BORFEO method improved emotion rating for the RFEs of anger, disgust, sadness, and surprise. No significant effect of BORFEO was observed for fearful and neutral expressions. Regarding happiness, emotion rating was higher for prototypical than BORFEO-generated expressions.




\subsection{Correlation between Py-Feat ratings and human ratings}

\begin{figure}[htpb]
  \centering
  \includegraphics[width=0.45\textwidth]{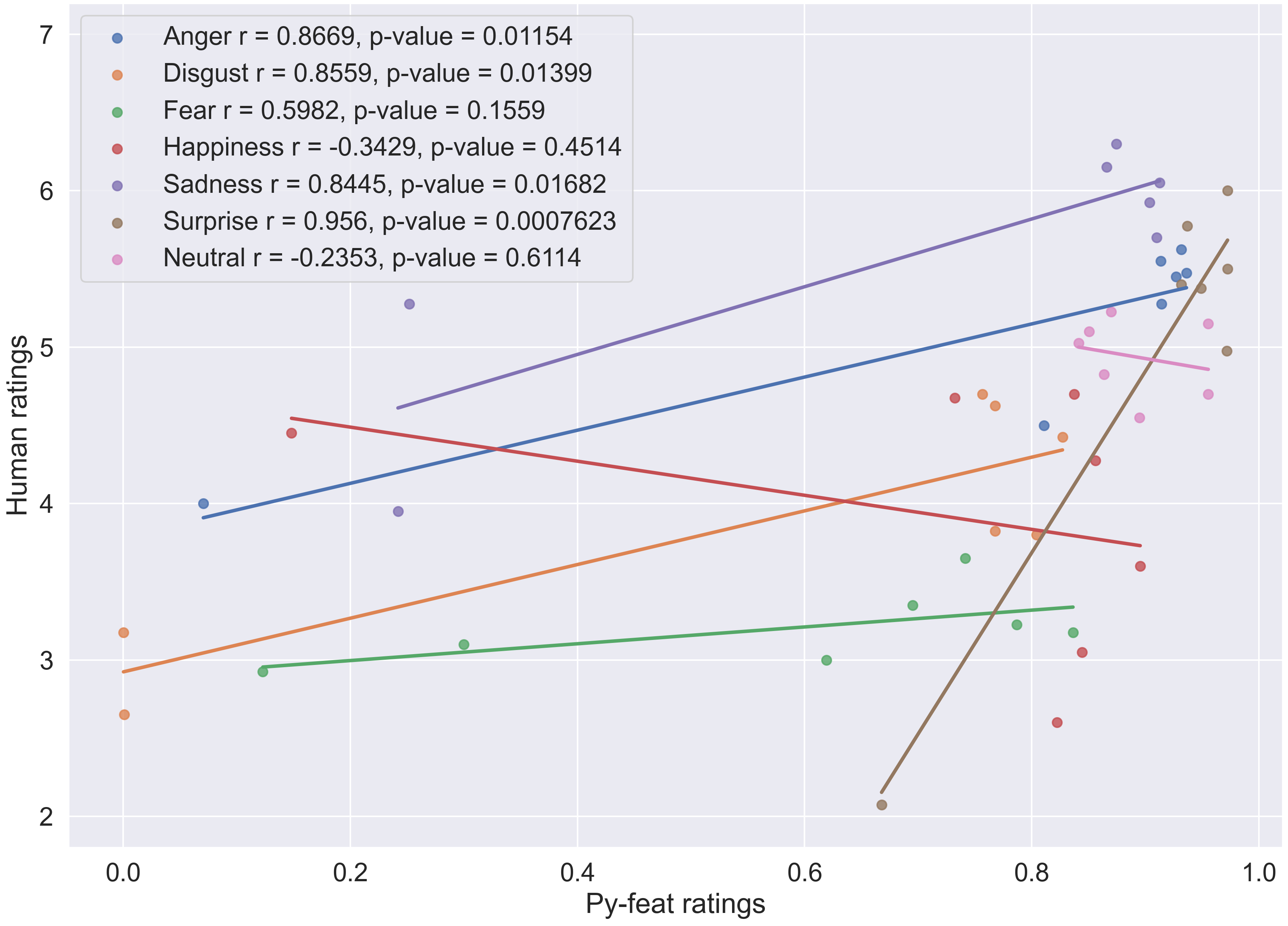}
  \caption{Scatter plots and regression lines for different types of stimuli.}
  \label{fig:correlation}
\end{figure}

Figure \ref{fig:correlation} shows the relationships (i.e., scatterplots and regression lines) between Py-Feat and human ratings for the stimulus photos of each target emotion. The human ratings indicated that Py-Feat successfully predicts human ratings for angry, sad, disgusted, and surprised expressions. However, Py-Feat did not successfully predict fearful, happy, or neutral expressions. Visual inspection of Figure \ref{fig:correlation} suggests a possible reason for the problems with these three expressions: positive correlations between Py-Feat and human ratings were observed for angry, disgusted, sad, and surprised expressions, while negative correlations were observed for fearful, happy, and neutral expressions.



Based on these results, we speculate that the poor performance of BORFEO concerning fearful, happy, and neutral expressions is related to disagreement between Py-Feat and human recognition of Nikola’s expressions. Py-Feat (ResMaskNet) is a machine learning algorithm trained on human ratings of human, rather than android, expressions. In addition, the results may have been influenced by cultural differences in emotion recognition \cite{becker2011intercultural} between the original raters and our participants. Prediction of how our participants would recognize Nikola’s facial emotions is an out-of-domain task for Py-Feat. We may be able to improve BORFEO performance by adjusting Py-Feat via domain adaptation \cite{daume2009frustratingly} or other transfer learning strategies. 


Humans often exhibit asymmetrical facial expressions \cite{lindell2018lateralization}. Although Nikola can produce asymmetrical expressions, we only considered symmetrical RFEs to simplify our analysis. Nikola may be able to produce more convincing expressions for human observers if we expand the parameter search range to include complex asymmetrical expressions.

\section{Conclusion}\label{sec:conclude}

In this work, we proposed and implemented the BORFEO method for an automatic RFE optimization task. The BORFEO method can be easily applied to any facial robot because it only requires a web camera, laptop, and means of transmitting robot control commands from the laptop; it does not require human experts with appropriate knowledge of facial expressions. We obtained a set of optimal control parameters for Nikola’s RFEs. The results of our human evaluation experiment showed that the BORFEO method improves Nikola’s RFEs for anger, disgust, sadness, and surprise compared with the RFEs generated by our previous method based on human coding \cite{sato_android_2022}. We also explored why the BORFEO method could not improve some expressions and offered suggestions to overcome this problem in future work.

\section{Acknowledgement}

This work was supported in part by JST, the establishment of university fellowships towards the creation of science technology innovation, Grant Number JPMJFS2123, and the JSPS Grants-in-Aid for Scientific Research (KAKENHI), Grant Number JP20H05957.

{\small
\bibliographystyle{./IEEEtran}
\bibliography{IEEEexample}
}

\end{document}